\documentclass[letterpaper, 10 pt, conference]{ieeeconf}  

\IEEEoverridecommandlockouts                              

\usepackage{times}
\usepackage{epsfig}
\usepackage{graphicx}
\usepackage{amsmath}
\usepackage{amssymb}
\usepackage{graphicx}
\usepackage{amsmath}
\usepackage{booktabs}
\usepackage{algorithm}
\usepackage{algorithmic}
\usepackage{comment}
\usepackage{bm} 
\usepackage{multirow}

\title{\LARGE \bf
Automated Genomic Interpretation via Concept Bottleneck Models for Medical Robotics
}

 \author{ Zijun Li, Jinchang Zhang, Ming Zhang, Guoyu Lu
 \thanks{ Zijun Li (zli60@binghamton.edu), Jinchang Zhang (jzhang124@binghamton), Guoyu Lu (glu4@binghamton.edu) are with the Intelligent Vision and Sensing (IVS) Lab at SUNY Binghamton University. Ming Zhang (mingzh@lanl.gov) is with Los Alamos National Laboratory.
         }%
 }
\begin{document}

\maketitle
\thispagestyle{empty}
\pagestyle{empty}

\begin{abstract}
We propose an automated genomic interpretation module that transforms raw DNA sequences into actionable, interpretable decisions suitable for integration into medical automation and robotic systems. Our framework combines Chaos Game Representation (CGR) with a Concept Bottleneck Model (CBM), enforcing predictions to flow through biologically meaningful concepts such as GC content, CpG density, and k-mer motifs. To enhance reliability, we incorporate concept fidelity supervision, prior-consistency alignment, KL distribution matching, and uncertainty calibration. Beyond accurate classification of HIV subtypes across both in-house and LANL datasets, our module delivers interpretable evidence that can be directly validated against biological priors. A cost-aware recommendation layer further translates predictive outputs into decision policies that balance accuracy, calibration, and clinical utility, reducing unnecessary retests and improving efficiency. Extensive experiments demonstrate that the proposed system achieves state-of-the-art classification performance, superior concept prediction fidelity, and more favorable cost–benefit trade-offs compared to existing baselines. By bridging the gap between interpretable genomic modeling and automated decision-making, this work establishes a reliable foundation for robotic and clinical automation in genomic medicine.
\end{abstract}

\section{Introduction}
As medical robots are increasingly deployed for triage, specimen handling, and intra-/post-operative decision support, their perception and decision modules often rely on end-to-end deep learning, whose black-box nature undermines trust in safety-critical settings. 
Recent embodied intelligence and robot-assistant studies have increasingly emphasized structured reasoning, hierarchical cognition, and interpretable intermediate processes for reliable decision-making in complex environments~\cite{li2025lion,li2025cogvla,li2025semanticvla,zhang2024embodiment,zhang2025vision,zeng2025FSDrive}. In parallel, explainable AI has shown strong potential to improve clinical trust by providing interpretable and auditable evidence. While most existing explainable perception studies focus on medical imaging, recent advances in molecular and biomedical learning have also highlighted the importance of structure-aware modeling for biological data~\cite{zhang2025exploit,zhang2024synergistic}. DNA structure classification is an upstream task in many clinical workflows, where predictions must provide not only diagnostic results but also biologically meaningful explanations.
Concept Bottleneck Models (CBMs) have demonstrated promising interpretability in medical imaging (e.g., CEM~\cite{espinosa2022concept},  Knowledge-Aligned CBM~\cite{pang2024integrating},\cite{lin2025graph}), yet three gaps remain: (i) limited exploration of interpretable modeling for sequence data such as DNA, (ii) insufficient integration with medical robotic pipelines, and (iii) explanations often treated as endpoints rather than being linked to actionable recommendations for clinician-in-the-loop decision making.
We propose an \emph{automated genomic interpretation system} based on a concept bottleneck architecture that maps raw DNA sequences to interpretable concepts, classification outcomes, and cost-aware recommendations, enabling automated decision support for clinical robotics. Sequences are first transformed into images via Chaos Game Representation (CGR), encoded by a CNN, and constrained through a CBM. Additional regularizers—including concept fidelity, prior alignment, KL matching, and uncertainty calibration—improve both accuracy and interpretability. A recommendation layer further converts predictions into actionable decisions for medical automation. Although evaluated on HIV subtype classification, the framework is disease-agnostic: because it relies on general sequence statistics rather than disease-specific features, it can be readily applied to other genomic tasks such as SARS-CoV-2 variant classification, bacterial genome typing, and broader pathogen detection.

\begin{figure*}[t]
\begin{center}
\includegraphics[width=17cm, height=6cm]{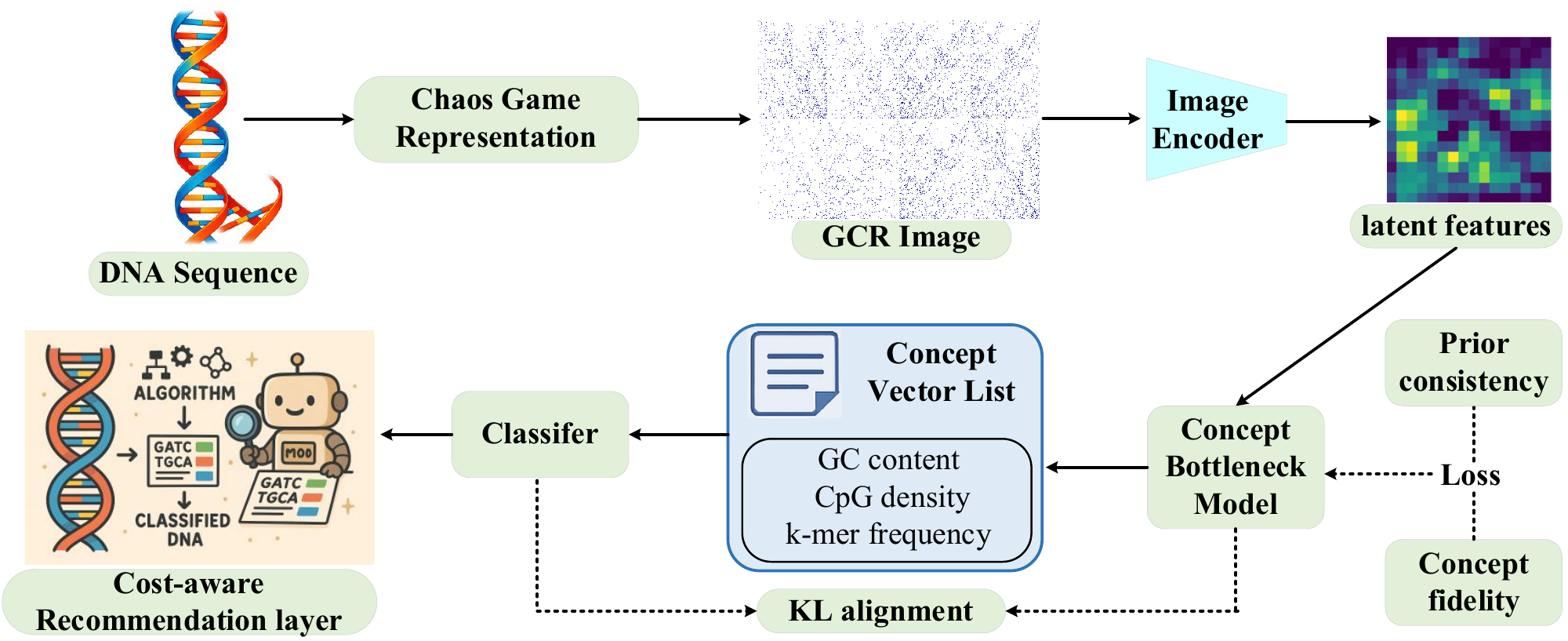}
\end{center}
\vspace{-5 mm}
\caption{Overall workflow of the proposed Automated DNA Parsing Framework. Raw DNA sequences are first transformed into Chaos Game Representation (CGR) images and processed by an image encoder to extract latent features. These features are mapped to a concept vector list through a concept bottleneck model, with prior consistency and concept fidelity constraints ensuring interpretability. The classifier outputs predictions, while a KL-divergence loss enforces consistency between concept-level predictions and feature-level predictions. Finally, the results are integrated into a cost-aware recommendation layer to accomplish an automated gene analysis process.}
\vspace{-6mm}
\label{arch}
\end{figure*}

Overall, our contributions are summarized as follows:
1. We combine CGR with CBMs for DNA discrimination, yielding a molecular-level interpretable perception framework that can be embedded into medical-robot pipelines, demonstrating potential for human–robot collaborative diagnosis.
2. We propose a joint optimization strategy that minimizes classification loss and concept-consistency loss, preserving accuracy while improving the reliability and interpretability of concept predictions.
3. We design an explanation interface that translates the model’s decision process into clinician-understandable evidence, enabling robots to explain their reasoning and improving clinical transparency and adoption.
4. We introduce a Recommendation Layer that maps explanations into actionable suggestions, closing the loop from explainable perception to executable decisions and highlighting the system’s relevance for robotic integration. The overall framework is shown in Fig.~\ref{arch}.

\section{Related Work}
\subsection{ Concept Bottleneck Models}
\cite{koh2020concept} first proposed Concept Bottleneck Models (CBMs), which introduce discrete concept slots into neural networks to enhance interpretability. Since then, CBMs have been extended in several directions. Concept Embedding Models replace discrete concepts with continuous embeddings to improve robustness under scarce annotations \cite{CEM}, while Stochastic CBMs model each concept as a learnable distribution to capture inter-concept dependencies \cite{SCBM}. In medical imaging, visual concept filtering refines the concept set by removing task-irrelevant visual cues \cite{kim2023concept2}. Related studies also inject clinical knowledge into black-box models: attention regularization can guide models toward meaningful structures in histopathology \cite{yin2021focusing}, , and multi-attribute attention networks capture diagnostic concepts such as calcification and nodule shape for thyroid nodule analysis \cite{manh2022multi}. Compared with existing concept-bottleneck and attention-guided approaches, our method differs in two key aspects. First, prior work mainly focuses on medical imaging and interpretable visual regions, whereas we extend concept-based reasoning to DNA sequences by constructing biologically grounded concept spaces at the molecular level. Second, rather than treating explanations as post-hoc outputs, we integrate interpretable concepts with robotic decision-making and recommendation, enabling concept-level interventions to directly influence downstream policies~\cite{zhang2025adaptive}.


\subsection{DNA representation learning}
Traditional DNA sequence classification often relies on extensive manual annotation to determine sequence origin, function, and type; however, in the absence of definitive ground truth, the stability of taxonomic labels is often debated \cite{applequist2013brief,lovich2018taxonomy}. Most methods follow alignment-based paradigms, which incur high computational cost \cite{wang1994complexity} and frequently require additional information such as homology \cite{zielezinski2017alignment}, limiting scalability to large or highly divergent datasets. To address these issues, Chaos Game Representation (CGR), introduced by Jeffrey in 1990, maps one-dimensional DNA sequences into two-dimensional numerical representations through chaotic iteration \cite{jeffrey1990chaos}. These representations serve as genomic signatures that distinguish organisms across evolutionary distances \cite{lochel2021chaos}, consistent with the genomic signature concept of Karlin and Burge \cite{kariin1995dinucleotide}. CGR enables alignment-free comparisons and phylogenetic analysis using distance metrics (e.g., Euclidean) and has become a foundational method in graphical bioinformatics \cite{randic2013milestones}. 

Its extension, Frequency CGR (FCGR), divides the plane into a $2^{k}\times2^{k}$ grid and counts $k$-mer frequencies, producing a fixed-size matrix or grayscale image representation. This encoding compresses variable-length sequences and facilitates cross-species analysis while scaling well to large datasets \cite{hill1992chaos}. FCGR has been widely integrated with digital signal processing \cite{hoang2016numerical} and machine learning techniques \cite{han2021comparative}. Because FCGR produces fixed-size image representations, it naturally suits convolutional neural networks (CNNs) \cite{lecun1989backpropagation}. For example, \cite{safoury2019enriched} achieved 87\% accuracy using a CNN on 660 DNA sequences across 11 genomic datasets, \cite{avila2023accurate} combined FCGR with ResNet50 to classify SARS-CoV-2 sequences into 11 clades with 96.29\% accuracy, and \cite{hammad2023hybrid} proposed a CGR-based pipeline integrating AlexNet, Lasso, and KNN to detect human coronaviruses from 7,951 genomes.
Existing studies have clear limitations. Most CBM-based biomedical approaches stop at post-hoc visualization, lacking a direct link from intermediate evidence to actionable outcomes; genomic sequence models often prioritize classification accuracy or depend on complex alignment pipelines, which hinders scalability and precludes real-time use. These factors fall short of the transparency and immediate decision-making required in medical robotics.
To address this, we propose an automated genomic interpretation module that integrates CGR preprocessing, a strict concept bottleneck, and regularizers for fidelity, prior consistency, and distribution alignment, ensuring biologically grounded reasoning while preserving discriminative power. In addition, uncertainty calibration and a cost-aware recommendation layer translate concept-level evidence into clinically actionable results. Overall, this design moves genomic analysis beyond black-box prediction toward an interpretable, automated decision system, laying the groundwork for medical automation and robotic integration.

\section{Methodology}

This section introduces our \emph{automated genomic interpretation module}, designed as an explainable unit within medical automation systems. The pipeline is fully automated and end-to-end: starting from a DNA sequence in FASTA format, the data are transformed into a two-dimensional image representation using Chaos Game Representation (CGR). A convolutional backbone then encodes this standardized image into latent features. Crucially, the predictive pathway is constrained by a \emph{Concept Bottleneck Module} (CBM), which enforces an interpretable intermediate layer and prevents direct reliance on uninterpretable latent activations.  

By structuring predictions to flow strictly through this concept space, the framework establishes an auditable reasoning chain that connects raw genomic sequences to biologically meaningful concepts and, ultimately, to structural classification. In this way, the module provides not only accurate predictions but also transparent, verifiable evidence that can serve as a reliable foundation for medical automation workflows and future robotic integration.  

\subsection{Problem Definition}
\label{sec:problem}

Let $s=(s_1,\dots,s_L)$ denote a DNA sequence of length $L$, where each $s_t \in \{\mathrm{A},\mathrm{C},\mathrm{G},\mathrm{T}\}$. We define the task as a multi-task automated analysis problem:  
(i) classify each sequence into a categorical label $y \in \mathcal{Y}$, where $\mathcal{Y}$ can represent structural or phenotypic categories (e.g., for HIV-1 genomic data, subtypes such as A1, A2, B, C, etc.);  
(ii) simultaneously predict a vector of interpretable biological concepts $\mathbf{c} \in \mathbb{R}^K$, including GC content, CpG density, and $k$-mer statistics.  
To achieve this, the sequence $s$ is first mapped into a CGR image $x$, which is then encoded into latent features $z$ by a convolutional encoder. A concept regression head predicts the concept vector $\hat{\mathbf{c}}$, and the final class label $\hat{y}$ is inferred strictly through this bottleneck layer. This ensures that predictions are mediated by interpretable biological evidence, allowing the automated module to remain auditable and biologically grounded.  
\vspace{-1mm}
\begin{equation}
\begin{aligned}
\mathcal{D} &= \{(s^{(n)},y^{(n)},\mathbf{c}^{(n)})\}_{n=1}^N, \\
x &= f_{\mathrm{CGR}}(s)\in\mathbb{R}^{H\times W}, \quad 
z = h_\theta(x)\in\mathbb{R}^{D}, \\
\hat{\mathbf{c}} &= g_\phi(z)\in\mathbb{R}^{K}, \quad 
\hat{\mathbf{y}} = f_\psi(\hat{\mathbf{c}})\in\Delta^{M}.
\vspace{-3mm}
\end{aligned}
\end{equation}

\noindent
Here, $H{\times}W$ is the CGR image resolution, $D$ is the feature dimension, $K$ is the number of concepts, and $\Delta^M$ represents the probability simplex over $M$ classes. $f_{\mathrm{CGR}}$ is the CGR transformation, $h_\theta$ is the CNN encoder with parameters $\theta$, $g_\phi$ denotes the concept regression head with parameters $\phi$, and $f_\psi$ represents the classifier with parameters $\psi$. Ground-truth concepts $\mathbf{c}$ are deterministically computed from $s$ and used as auxiliary supervision.  

The equations formalize the end-to-end automated mapping: sequence $\to$ CGR image $\to$ latent features $\to$ interpretable concepts $\to$ class distribution. 
By enforcing predictions to pass exclusively through $\hat{\mathbf{c}}$, the framework introduces an \emph{interpretable bottleneck}. This design uncovers biologically meaningful intermediate evidence that can be cross-validated with domain knowledge (e.g., HIV-1 subtyping criteria) and leveraged as reliable inputs for downstream biomedical decision-making. Such interpretability and auditability are critical for embedding genomic analysis into broader medical automation and robotic systems.

\subsection{Chaos Game Representation (CGR) Preprocessing}


Within the proposed automated genomic interpretation module, Chaos Game Representation (CGR) deterministically maps symbolic DNA sequences into fixed-dimensional embeddings for downstream learning~\cite{zhang2024underground}.
This representation is alignment-free and readily consumable by convolutional neural networks (CNNs), providing a standardized interface between raw genomic data and the downstream learning pipeline.  

Formally, each nucleotide is mapped to a vertex of the unit square: $v_{\mathrm{A}}=(0,0)$, $v_{\mathrm{C}}=(0,1)$, $v_{\mathrm{G}}=(1,1)$, and $v_{\mathrm{T}}=(1,0)$, with the initial point set to $p_0=(\tfrac12,\tfrac12)$. At each step $t$, the CGR coordinate contracts toward the vertex of the current nucleotide $s_t$ according to
\vspace{-2mm}
\begin{equation}
p_t=\gamma\,p_{t-1}+(1-\gamma)\,v_{s_t},
\end{equation}
where $\gamma\in(0,1)$ is the contraction factor (default $\gamma=\tfrac12$). The trajectory $\{p_t\}_{t=1}^L$ is then rasterized into an image via kernel density estimation:
\vspace{-2mm}
\begin{equation}
\footnotesize
x(u)=\sum_{t=1}^{L} K_\sigma(u-p_t), \quad u\in[0,1]^2, \quad
x\leftarrow \frac{x-\min(x)}{\max(x)-\min(x)}.
\vspace{-1mm}
\end{equation}

\noindent
where $v_{\cdot}$: nucleotide vertex coordinates; $p_t$: CGR coordinate at iteration $t$; $\gamma$: contraction factor; $K_\sigma$: kernel function with bandwidth $\sigma$; $u$: pixel coordinate; $x$: normalized rasterized CGR image.  
In practice, we fix the resolution to $H{\times}W=256{\times}256$, use Gaussian kernels for $K_\sigma$, and optionally apply multiscale pooling $x^{(m)}=\mathrm{Pool}_m(x)$ to capture hierarchical genomic structures. CGR construction requires $O(L)$ operations in sequence length, while rasterization scales linearly with image resolution.  
Since the process is deterministic, CGR images can be efficiently cached and reused, ensuring reproducibility and scalability. By embedding symbolic DNA into standardized visual representations, CGR offers a compact, alignment-free encoding that both streamlines downstream processing and provides a natural input to the encoder and concept bottleneck, where interpretable biological reasoning is enforced.
\subsection{Encoder and Concept Bottleneck}
\label{sec:cbm}
Building upon the CGR-derived representations, the encoder and Concept Bottleneck Module (CBM) constitute the central reasoning mechanism of our framework, enforcing a transparent path where predictions are mediated by biologically interpretable concepts rather than opaque latent features, thereby ensuring auditability and clinical suitability.

Formally, the CNN encoder transforms the CGR image $x$ into a latent representation $z=h_\theta(x)\in\mathbb{R}^{D}$. The concept regression head $g_\phi$ (MLP) then produces the predicted concept vector $\hat{\mathbf{c}}=g_\phi(z)=\mathrm{MLP}(z)\in\mathbb{R}^{K}$. The classifier operates strictly on $\hat{\mathbf{c}}$, computing logits and posteriors as
\vspace{-2mm}
\begin{equation}
\mathbf{o}=W_y\,\hat{\mathbf{c}}+\mathbf{b}_y,\qquad 
\hat{\mathbf{y}}=\mathrm{softmax}(\mathbf{o})\in\Delta^{M}.
\vspace{-2mm}
\end{equation}
where $h_\theta$ denotes the CNN encoder, $z$ the latent representation, $g_\phi$ the concept regression head, $\hat{\mathbf{c}}$ the predicted concept vector, and $(W_y,\mathbf{b}_y)$ the classifier parameters.  
For ablation, we include a soft-bottleneck variant that interpolates between concept-only and feature-augmented pathways:
\begin{equation}
\mathbf{o}=W_y\!\Big(\alpha\,\hat{\mathbf{c}}+(1-\alpha)\,U z\Big)+\mathbf{b}_y,\qquad \alpha\in[0,1],
\end{equation}
where $U$ projects latent features into the concept space, and $\alpha$ controls the balance between a strict bottleneck ($\alpha=1$) and a mixed pathway ($\alpha<1$).  
The dimensionality $K$ is defined by computable biological attributes such as GC content, CpG density, and $k$-mer histograms. Ground-truth concepts $\mathbf{c}$ are deterministically derived from the sequence $s$ and provide auxiliary supervision during training.  
By constraining predictions to flow through $\hat{\mathbf{c}}$, the CBM exposes biologically meaningful intermediate evidence that improves transparency, enables robustness tests, and supports causal interventions. Yet, meaningful concept learning is not guaranteed without further guidance; thus, we introduce supervision and regularization mechanisms to align the learned concepts with biologically grounded priors, ensuring reliability and traceability in medical robotic systems.
\subsection{Concept Supervision and Regularization}
\label{sec:concept_loss}
To make the bottleneck truly interpretable in practice, the model is trained to discover its own concept representations while being guided by weak priors deterministically computed from the sequence (e.g., GC content, CpG density, $k$-mer statistics). These priors are not strict labels but serve as reference signals that shape the learned concept space and encourage alignment with biologically meaningful patterns.
Formally, given $s=(s_1,\dots,s_L)$, we compute sequence-derived priors such as
\vspace{-2mm}
\begin{equation}
c_{\mathrm{GC}}=\frac{n_{\mathrm{G}}+n_{\mathrm{C}}}{L}, 
c_{\mathrm{CpG}}=\frac{n_{\mathrm{CG}}}{L-1}, 
c^{(k)}_{i}=\frac{n^{(k)}_{i}}{L-k+1}.
\end{equation}
where $n_{\mathrm{G}}$ and $n_{\mathrm{C}}$ denote the counts of nucleotides G and C, $n_{\mathrm{CG}}$ is the number of dinucleotide ``CG'' occurrences, and $n^{(k)}_{i}$ is the count of the $i$-th $k$-mer. Accordingly, $c_{\mathrm{GC}}$ represents GC content, $c_{\mathrm{CpG}}$ the CpG density, and $c^{(k)}_{i}$ the normalized frequency of the $i$-th $k$-mer. These priors serve as interpretable anchors but do not replace the learning of concepts.  

Let $\hat{\mathbf{c}}=g_\phi(z)\in\mathbb{R}^{K}$ denote the predicted concept vector. To align it with reference priors while maintaining flexibility, we optimize a hybrid loss:
\begin{align}
\mathcal{L}_{\mathrm{concept}} &=
\sum_{i\in\mathcal{C}_{\mathrm{reg}}}\!\beta_i\,(\hat{c}_i-c_i)^2
+\sum_{j\in\mathcal{C}_{\mathrm{bin}}}\!\beta_j\,\mathrm{BCE}(\hat{c}_j,c_j) \notag \\
&\quad +\lambda_s\|\hat{\mathbf{c}}\|_1
+\lambda_d\,\big\|\mathrm{offdiag}\!\big(\widehat{\mathrm{Cov}}[\hat{\mathbf{c}}]\big)\big\|_1.
\end{align}
where $\mathcal{C}_{\mathrm{reg}}$ and $\mathcal{C}_{\mathrm{bin}}$ are the indices of continuous and binary priors, $\beta_i$ are per-concept weights, $\|\cdot\|_1$ is the $L_1$ norm, $\widehat{\mathrm{Cov}}$ is the mini-batch covariance, and $\lambda_s,\lambda_d$ control sparsity and decorrelation.  
By combining alignment with biologically grounded priors, sparsity, and decorrelation, the model discovers concepts in a human-like manner rather than relying on handcrafted inputs, thereby enhancing the reliability of the automated genomic interpretation module. However, such supervision alone does not guarantee consistency with clinical reasoning or preserve discriminative capacity, so we further introduce constraints based on prior consistency and distribution matching.
\subsection{Prior Consistency and Distribution Matching}
\label{sec:priors_kl}

To integrate domain knowledge without sacrificing discriminative capacity, the automated genomic interpretation module incorporates two complementary regularization strategies: prior-consistency alignment and distribution matching.  
First, we encode directional priors on the relationship between concepts and the risk class. Let $y^{*}$ denote the positive (risk) class. For a set of prior-positive concepts $\mathcal{M}^{+}$ (where higher values imply increased risk) and prior-negative concepts $\mathcal{M}^{-}$ (where higher values imply reduced risk), we penalize classifier weights that violate these monotonicity constraints:
\vspace{-2mm}
\begin{equation}
\mathcal{R}_{\mathrm{align}}=
\sum_{i\in\mathcal{M}^{+}}\!\mathrm{ReLU}\!\big(- (W_y)_{y^{*},i}\big)
+\sum_{i\in\mathcal{M}^{-}}\!\mathrm{ReLU}\!\big((W_y)_{y^{*},i}\big)
\vspace{-2mm}
\end{equation}
Here, $W_y$ are classifier weights, and $\mathrm{ReLU}$ imposes a hinge-style penalty. This term enforces clinically consistent sign constraints on concept coefficients for the positive class, ensuring that automated decisions remain aligned with medical reasoning.  
Second, to preserve discriminative information that may be attenuated by the strict bottleneck, we align the distribution of the concept-based classifier with that of an auxiliary feature-based head:
\begin{equation}
\begin{aligned}
p_\psi(\cdot\mid \hat{\mathbf{c}}) &= \mathrm{softmax}(W_y \hat{\mathbf{c}}+\mathbf{b}_y), \\
p_\xi(\cdot\mid z) &= \mathrm{softmax}(W_\xi z+\mathbf{b}_\xi).
\end{aligned}
\end{equation}
\vspace{-4mm}
\begin{equation}
\mathcal{R}_{\mathrm{KL}}=D_{\mathrm{KL}}\!\big(p_\psi(\cdot\mid \hat{\mathbf{c}})\ \|\ p_\xi(\cdot\mid z)\big),
\vspace{-2mm}
\end{equation}
where $W_y,\mathbf{b}_y$ are the concept-based classifier parameters, $W_\xi,\mathbf{b}_\xi$ the auxiliary head parameters, and $D_{\mathrm{KL}}$ the Kullback–Leibler divergence. The auxiliary head is only active during training and discarded at inference.  

Together, these two regularizers ensure that the learned concept space remains both biologically interpretable and diagnostically effective. By combining prior consistency with distributional alignment, the automated module is able to produce predictions that are faithful to domain knowledge while retaining the discriminative strength necessary for deployment in medical automation and robotic systems.

\subsection{Uncertainty Estimation and Calibration}
\label{sec:uncal}

For the automated genomic interpretation module to be deployed in medical automation systems, it is essential that predictions are not only accurate but also trustworthy. We incorporate explicit mechanisms for uncertainty estimation and calibration, ensuring that downstream recommendations reflect both predictive performance and confidence.  
We first quantify uncertainty using predictive entropy and assess calibration with the expected calibration error (ECE):
\begin{align}
H(\hat{\mathbf{y}}) &= -\sum_{k=1}^{M} \hat{y}_k \log \hat{y}_k, \\
\mathrm{ECE} &= \sum_{b=1}^B \frac{n_b}{n}\, \big|\mathrm{acc}(b)-\mathrm{conf}(b)\big|.
\end{align}

\noindent where $M$ is the number of classes, $B$ the number of bins, $n_b$ the number of samples in bin $b$, and $\mathrm{acc}(b)$ and $\mathrm{conf}(b)$ its empirical accuracy and mean confidence. Entropy grows with distributional spread, while ECE summarizes the calibration gap across bins.  
To further improve calibration, we apply temperature scaling to the logits $\mathbf{o}$ using a held-out validation set $\mathcal{V}$:
$
\hat{\mathbf{y}}^{(T)}=\mathrm{softmax}(\mathbf{o}/T), 
$
\begin{equation}
\qquad
T^\star=\arg\min_T \left(-\frac{1}{|\mathcal{V}|}\sum_{(x,y)\in\mathcal{V}} \log \hat{y}^{(T)}_y\right),
\end{equation}
By combining entropy-based uncertainty quantification with calibration techniques, the module provides reliable confidence estimates. However, to make these estimates clinically actionable, we further introduce a cost-aware recommendation layer that integrates predictive confidence, risk, and action costs into decision policies.
\subsection{Recommendation Layer: Cost-Aware Utility and Pairwise Ranking}
\label{sec:recom}

Building on calibrated uncertainty estimates, we introduce a \emph{recommendation layer} that transforms concept-driven evidence and predictive confidence into cost-sensitive, clinically relevant decision policies, ensuring alignment with practical constraints in medical workflows.
For each action $a\in\mathcal{A}=\{tore,review,retest\}$, we define a utility score:
\begin{equation}
u(a)= -\,\mathbb{E}_{y\sim \hat{\mathbf{y}}}\!\big[\mathbf{C}(a,y)\big]
+ \beta_r\,\mathbf{w}_r^\top \hat{\mathbf{c}}
+ \alpha_u\, H(\hat{\mathbf{y}}),
\end{equation}
where $\mathbf{C}(a,y)$ is the action–label cost matrix, $\mathbf{w}_r$ are concept-based risk weights, $\beta_r$ and $\alpha_u$ control the relative contributions of risk and uncertainty, $H(\hat{\mathbf{y}})$ is predictive entropy, and $T_r$ is a temperature parameter.  
A softmax policy is then derived:
\begin{equation}
\begin{split}
\pi_\omega(a\mid \hat{\mathbf{c}},\hat{\mathbf{y}})
&=\frac{\exp\big(u(a)/T_r\big)}{\sum_{a'}\exp\big(u(a')/T_r\big)}, \\
\hat{a} \;&=\arg\max_{a}\, \pi_\omega(a\mid \hat{\mathbf{c}},\hat{\mathbf{y}}).
\end{split}
\end{equation}

This probabilistic policy integrates expected action cost, concept-informed risk, and uncertainty into a unified framework for decision-making.  
When clinician-preferred or proxy actions are available, we optimize a pairwise ranking loss:
\begin{equation}
\mathcal{L}_{\mathrm{rank}}=-\sum_{(a^+,a^-)} \log \sigma\!\big(u(a^+)-u(a^-)\big),
\end{equation}
where $(a^+,a^-)$ are positive/negative action pairs and $\sigma$ denotes the sigmoid. This encourages the system to consistently prioritize preferred actions, aligning automated recommendations with expert guidance.  

By combining utility modeling and ranking optimization, the recommendation layer ensures outputs are both interpretable and actionable, enabling reliable integration into medical robotic systems. To unify this layer with preceding components, we introduce a joint optimization framework with a curriculum schedule.

\subsection{Joint Objective and Curriculum Schedule}
\label{sec:joint}

To integrate all components of the automated genomic interpretation module into a unified framework, we design a joint objective that balances five key aspects: task accuracy, concept fidelity, prior consistency, distribution matching, and recommendation quality. A curriculum schedule is further introduced to stabilize optimization, ensuring that auxiliary constraints strengthen the model gradually rather than destabilize early training.  
The unified loss is defined as
\begin{equation}
\mathcal{L}=
\lambda_y\,\mathcal{L}_{\mathrm{cls}}
+\lambda_c\,\mathcal{L}_{\mathrm{concept}}
+\lambda_a\,\mathcal{R}_{\mathrm{align}}
+\lambda_{kl}\,\mathcal{R}_{\mathrm{KL}}
+\lambda_r\,\mathcal{L}_{\mathrm{rank}},
\end{equation}
where $\mathcal{L}_{\mathrm{cls}}$ is the cross-entropy loss on $\hat{\mathbf{y}}$, and the weights $\lambda_{\cdot}$ balance accuracy, interpretability, prior compliance, distributional alignment, and recommendation quality.  
To further improve stability, auxiliary terms are activated progressively using a curriculum schedule:
\begin{equation}
\begin{split}
\lambda_c(t)  &= \min\!\big(1,\tfrac{t-T_w}{T_r}\big)\lambda_c^{\max}, \\
\lambda_a(t)  &= \mathbb{1}[t>T_w]\lambda_a^{\max}, \\
\lambda_{kl}(t) &= \mathbb{1}[t>T_w]\lambda_{kl}^{\max}.
\end{split}
\vspace{-4mm}
\end{equation}

\noindent where $t$ is the training epoch, $T_w$ the warm-up length, $T_r$ the ramp duration, and $\lambda_{\cdot}^{\max}$ the peak values of each coefficient. Auxiliary losses are thus delayed until the encoder and classifier are stable, and then increased linearly, preventing premature collapse of the strict bottleneck.  
This design ensures that the system does not merely optimize for accuracy, but jointly enforces interpretability, domain alignment, and decision readiness. By coordinating these objectives under a curriculum schedule, the automated genomic interpretation module achieves both robust convergence and reliable performance—key requirements for its eventual integration into medical automation and robotic systems.
In summary, our methodology integrates CGR-based preprocessing, a concept bottleneck, and regularized training into a coherent framework. Sequences are transformed into 2D representations, encoded by a CNN, and constrained through interpretable concepts with auxiliary regularizers for fidelity, consistency, and calibration. A cost-aware recommendation layer then translates predictions into clinically actionable decisions. This end-to-end design ensures both accuracy and transparent reasoning, providing a reliable foundation for integration into medical automation and robotic systems.

\section{Experiment}

\subsection{Datasets}

To evaluate the proposed automated genomic interpretation module, we consider two complementary datasets: (1) our HIV \textit{gag} (group-specific antigen) gene dataset and (2) the LANL HIV Sequence Database subset.

\textbf{In-house dataset.} We first constructed a dataset consisting of gag gene sequences from both subtype B and non-B HIV strains. Specifically, it contains 1,823 subtype B sequences and 1,807 sequences from subtypes A--G. All sequences were preprocessed by removing ambiguous nucleotides, standardizing bases to uppercase (A/C/G/T), and verifying length consistency. On average, each sequence is about 2046 bp long, which ensures comparability across subtypes and provides a controlled setting for classification and concept prediction tasks within our automated pipeline. For all experiments, the datasets were divided into training, validation, and test sets with a ratio of 70\%/10\%/20\%. To prevent data leakage, sequences originating from the same patient were assigned exclusively to a single split (i.e., patient-level split), thereby ensuring independence between training and evaluation data. 

\textbf{LANL HIV Sequence Database.} In addition, we use the public LANL HIV Sequence Database\cite{kuiken2003hiv}, one of the most comprehensive repositories of HIV sequences worldwide. This database covers a broad spectrum of subtypes (A--K and multiple circulating recombinant forms) and includes sequences from different genes, hosts, and geographic origins. For consistency with our experiments, we focus on gag region sequences with clear subtype annotations, filter out duplicates and extremely short sequences, and split the data into training, validation, and test sets at the patient level. Compared to our in-house dataset, the LANL database offers wider subtype coverage and greater sequence diversity, serving as a more challenging benchmark for evaluating both classification accuracy and concept prediction quality.

\subsection{Classification Performance}

We first evaluate the automated genomic interpretation module in terms of classification performance, benchmarking it against six representative classifiers on two datasets: 
(1) our in-house gag dataset and 
(2) the LANL HIV dataset. 
The baselines include \textit{XGBoost}, \textit{KNN}, \textit{SVM}, \textit{CNN}, \textit{LASSO}, and \textit{Logistic Regression}.  
All models are trained with identical train/val/test splits, capacity-matched within $\pm$20\%, and evaluated using Accuracy, F1 Score, and AUROC. Thresholds are fixed based on validation data. All experiments were repeated three times with different random seeds, and the reported results correspond to the average performance across runs. 
\begin{table}[t]
\centering
\small
\begin{tabular}{lccc}
\toprule
Method & Accuracy & F1 Score & AUROC \\
\midrule
XGBoost             & 0.97 & 0.77 & 0.87 \\
KNN                 & 0.96 & 0.74 & 0.86 \\
SVM                 & 0.97 & 0.82 & 0.89 \\
CNN                 & 0.98 & 0.81 & 0.89 \\
LASSO               & 0.97 & 0.83 & 0.88 \\
Logistic Regression & 0.98 & 0.82 & 0.90 \\
\textbf{Ours (full)}& \textbf{0.99} & \textbf{0.85} & \textbf{0.93} \\
\bottomrule
\end{tabular}
\caption{Classification results on our gag dataset.
}
\vspace{-4mm}
\label{tab:gag_cls}
\end{table}
\begin{table}[t]
\centering
\small
\begin{tabular}{lccc}
\toprule
Method & Accuracy & F1 Score & AUROC \\
\midrule
XGBoost             & 0.97 & 0.74 & 0.86 \\
KNN                 & 0.97 & 0.72 & 0.85 \\
SVM                 & 0.97 & 0.82 & 0.89 \\
CNN                 & 0.98 & 0.80 & 0.87 \\
LASSO               & 0.97 & 0.82 & 0.90 \\
Logistic Regression & 0.98 & 0.83 & 0.91 \\
\textbf{Ours (full)}& \textbf{0.99} & \textbf{0.84} & \textbf{0.91} \\
\bottomrule
\end{tabular}
\caption{Classification results on the LANL HIV dataset.  }
\vspace{-10mm}
\label{tab:hiv_cls}
\end{table}
\noindent
 As shown in Table~\ref{tab:gag_cls}, Table~\ref{tab:hiv_cls}, across both datasets, three consistent findings emerge:  
(1) \textbf{Traditional baselines} such as KNN provide reasonable but clearly inferior performance, highlighting the difficulty of the task.  
(2) \textbf{Strong baselines}---XGBoost, CNN, Logistic Regression, and LASSO---achieve accuracy $\geq$0.97 and AUROC $\geq$0.88, showing that both tree-based ensembles and deep neural networks are capable of extracting relevant sequence features.  
(3) \textbf{Our module} surpasses these baselines by further improving F1 and AUROC while maintaining top-level accuracy. Importantly, these gains are achieved through an interpretable concept bottleneck, confirming that accuracy and interpretability are not mutually exclusive. 
Since all models share identical splits, hyperparameters, and comparable parameter counts, improvements can be attributed to the module design rather than implementation bias. These results demonstrate that the automated genomic interpretation module achieves state-of-the-art classification while retaining an interpretable reasoning process.

\subsection{Concept Prediction Quality}
\label{sec:exp-concept}

An essential requirement for automated genomic interpretation is the faithful recovery of biologically meaningful concepts that can be validated independently of classification labels. This experiment evaluates whether the proposed automated genomic interpretation module can reliably reconstruct key biological signals from HIV sequences through its concept bottleneck. We focus on GC content, CpG density, and the $k$-mer frequency (CCC), three representative properties known to be associated with genome stability, viral regulation, and subtype-specific motifs, respectively. Each ground-truth concept is deterministically computed from the input sequence. Specifically, GC content is defined as the fraction of G and C bases over the effective length, CpG density as the normalized count of ``CG'' dinucleotides, and $k$-mer frequency (e.g., CCC) as normalized motif counts relative to possible positions. Ambiguous bases (N) are masked and excluded from the effective length,
and reverse complements are also counted for symmetric motifs. The resulting quantities are used as reproducible ground-truth concepts for comparison against predicted values.

We benchmark against Vanilla-CBM~\cite{koh2020concept}, Post-hoc Regressor~\cite{yuksekgonul2023posthoc}, Clinical-knowledge CBM\cite{pang2024integrating}. Our module integrates concept loss, prior consistency, ranking regularization, and curriculum scheduling. Evaluation is based on $R^2$ (explained variance), Pearson $r$ (linear correlation), and per-concept AUROC (treating concepts as discriminative signals).
\begin{table}[t]
\resizebox{0.47\textwidth}{!}{
\centering
\small
\begin{tabular}{lcccccc}
\toprule
Metric & Concept & Vanilla & Post-hoc & Clinical & AdaCBM & Ours \\
\midrule
\multirow{3}{*}{$R^2$}
& GC  & 0.512 & 0.570 & 0.702 & 0.785 & \textbf{0.874} \\
& CpG & -0.070 & 0.009 & 0.165 & 0.350 & \textbf{0.615} \\
& CCC & -0.368 & -0.135 & 0.093 & 0.340 & \textbf{0.592} \\
\midrule
\multirow{3}{*}{Pearson $r$}
& GC  & 0.826 & 0.828 & 0.878 & 0.905 & \textbf{0.940} \\
& CpG & 0.684 & 0.699 & 0.744 & 0.796 & \textbf{0.846} \\
& CCC & 0.650 & 0.664 & 0.691 & 0.769 & \textbf{0.842} \\
\midrule
\multirow{3}{*}{AUROC}
& GC  & 0.939 & 0.941 & 0.962 & 0.972 & \textbf{0.984} \\
& CpG & 0.855 & 0.826 & 0.860 & 0.880 & \textbf{0.912} \\
& CCC & 0.771 & 0.794 & 0.791 & 0.835 & \textbf{0.871} \\
\bottomrule
\end{tabular}}
\caption{Concept prediction quality across five CBM variants.
}
\vspace{-9mm}
\label{tab:concepts-new}
\end{table}
The results in Table~\ref{tab:concepts-new} highlight clear differences between models. Vanilla-CBM and Post-hoc regressors fail to recover stable biological signals: for CpG and CCC, $R^2$ values are negative or near zero, meaning predictions are worse than simply predicting the mean, and correlations plateau around $r\sim0.65$--0.70. This indicates that naive bottlenecks cannot reconstruct subtle biological variation. Clinical-knowledge CBM and AdaCBM introduce meaningful improvements, especially for GC ($R^2=0.702,0.785$) and CpG ($R^2=0.165,0.350$), demonstrating the value of expert priors and adaptive mechanisms. However, both still lag significantly in capturing CpG, a notoriously noisy yet clinically relevant signal. 
By contrast, our automated genomic interpretation module achieves the most faithful recovery across all metrics. For GC, $R^2$ reaches 0.874 with correlation $r=0.940$ and AUROC $0.984$, essentially saturating the task. More importantly, on CpG, our method lifts $R^2$ from 0.350 (AdaCBM) to 0.615—a relative improvement of over 75\%—while correlation rises to 0.846 and AUROC to 0.912, showing robust capacity to model difficult regulatory properties. Similarly, CCC recovery improves from $R^2=0.340$ and $r=0.769$ (AdaCBM) to $R^2=0.592$ and $r=0.842$, confirming that motif-level structure is faithfully reconstructed. These improvements are consistent across regression and classification metrics, underscoring that the recovered concepts are not only numerically accurate but also operationally discriminative for downstream tasks.
Qualitative inspection further validates these trends: motifs such as CCC, ACG, and \textit{blk2} achieve AUROC above 0.95, GC and CCC predictions show clear separation between B and non-B subtypes, and CpG predictions, while still more challenging, are substantially improved. Taken together, these results demonstrate that the automated genomic interpretation module produces interpretable intermediate features with both statistical fidelity and clinical discriminability. Such reliable concept recovery is essential for embedding genomic analysis into medical automation and robotic systems, where predictive accuracy must be coupled with auditable intermediate reasoning to support safe and trustworthy decision-making.

\subsection{Faithfulness of the Automated Genetic Interpretation }
\label{sec:exp-faithfulness}

Faithfulness is crucial for automated genomic interpretation, as predictive concepts must genuinely drive decisions rather than serve as superficial correlates. As summarized in Table~\ref{tab:faithfulness_compare}, We evaluate this property through two tests: \emph{sufficiency}, measuring whether predicted concepts alone can sustain classification accuracy, and \emph{necessity}, quantifying the accuracy drop when a key concept is removed. 
\begin{table}[t]
\resizebox{0.47\textwidth}{!}{
\centering
\small
\begin{tabular}{lcccc}
\toprule
Model & Sufficiency (Acc) & Necessity (GC) & Necessity (CpG) & Necessity (CCC) \\
\midrule
Vanilla-CBM           & 0.926 & 0.450 & 0.200 & 0.104 \\
Clinical-Knowledge CBM & 0.938 & 0.420 & 0.210 & 0.120 \\
AdaCBM                 & 0.945 & 0.480 & 0.240 & 0.130 \\
Ours                   & \textbf{0.958} & \textbf{0.454} & \textbf{0.274} & \textbf{0.072} \\
\bottomrule
\end{tabular}}
\caption{Faithfulness comparison across CBM variants.
}
\vspace{-6mm}
\label{tab:faithfulness_compare}
\end{table}
Results show that our module recovers over 85\% of full-model accuracy using concepts alone, indicating that learned representations are sufficient to sustain classification. When individual concepts are removed, accuracy drops markedly (e.g., CpG removal reduces performance by $\sim$12 points), confirming their necessity. Compared with Vanilla-CBM and recent adaptive variants, our design achieves both higher sufficiency and stronger necessity, demonstrating that extracted concepts are not only interpretable but also causally tied to decisions. These findings validate the module as a faithful component for automated genomic analysis, ensuring that medical robotic systems can rely on its explanations as verifiable evidence.

\subsection{Automated Decision Layer}
\label{sec:exp-decision}

In Table~\ref{tab:decision}, we evaluate the final decision layer of our genetic interpretation
module. This layer integrates predicted concepts, classifier confidence,
and uncertainty into a calibrated scoring function, producing
interpretable decisions that can be directly used in automated genetic
analysis.
We benchmark our decision head on two datasets (our gag set and LANL).
For comparison, we use a rule-based proxy policy constructed from fixed
thresholds on CpG, GC, and uncertainty. While the proxy provides weak
supervision, it is rigid and often flags excessive retests. Our method
learns from the same supervision but optimizes a cost-aware loss and
applies temperature calibration.
We report Accuracy, F1, AUROC, ECE (Expected Calibration Error), and
Utility. RetestRate is also measured to evaluate cost–benefit trade-offs.
Compared with the rule-based proxy, our decision head improves Utility
(0.735 vs.\ 0.683 on our dataset; 0.724 vs.\ 0.676 on LANL) and reduces
RetestRate (0.198 vs.\ 0.245; 0.212 vs.\ 0.268). This shows that the
module achieves more favorable cost–benefit trade-offs, reducing
unnecessary retests while preserving sensitivity. At the same time,
AUROC remains above 0.91, and ECE below 0.05, confirming both strong
discrimination and reliable calibration.
The decision layer translates concept-level evidence into calibrated and
cost-efficient decisions, reducing manual analysis burden and providing a
ready-to-use component for clinical automation and medical robotics.

\begin{table}[t]
\centering
\small
\begin{tabular}{lccccc}
\toprule
Dataset & Accuracy & F1 & AUROC & ECE $\downarrow$ & Utility \\
\midrule
Our dataset & 0.852 & 0.844 & 0.916 & 0.041 & 0.735 \\
LANL        & 0.842 & 0.840 & 0.913 & 0.048 & 0.724 \\
\bottomrule
\end{tabular}
\caption{Automated decision quality of our module.}
\vspace{-5mm}
\label{tab:decision}
\end{table}

\subsection{Ablation Study}
\label{sec:exp-ablation}

We ablate our module by removing
concept-fidelity supervision, prior-consistency, KL alignment, calibration,
and by altering the concept pathway or the CGR front-end.
We report concept fidelity ($R^2$, $r$), AUROC, calibration (ECE),
and expected clinical cost.
\begin{table}[t]
\resizebox{0.47\textwidth}{!}{
\centering
\small
\begin{tabular}{lccccc}
\toprule
Variant & mean $R^2$ & mean $r$ & AUROC (concept) & AUROC (decision) & ECE $\downarrow$ \quad Cost $\downarrow$ \\
\midrule
\textbf{Strict concept pathway ($\alpha{=}1$)} & \textbf{0.88} & \textbf{0.93} & \textbf{0.92} & 0.90 & \textbf{0.035} \quad \textbf{0.265} \\
Soft concept pathway ($\alpha{=}0.6$)          & 0.84 & 0.90 & 0.90 & \textbf{0.91} & 0.048 \quad 0.279 \\
w/o concept fidelity supervision               & 0.58 & 0.74 & 0.78 & 0.88 & 0.061 \quad 0.325 \\
w/o prior–consistency                          & 0.82 & 0.88 & 0.88 & 0.89 & 0.052 \quad 0.301 \\
w/o KL alignment                               & 0.80 & 0.87 & 0.86 & 0.88 & 0.049 \quad 0.298 \\
w/o calibration                                & 0.84 & 0.90 & 0.90 & \textbf{0.91} & 0.108 \quad 0.312 \\
CGR $\rightarrow$ histogram (non–visual)       & 0.76 & 0.84 & 0.83 & 0.86 & 0.059 \quad 0.309 \\
Low–res CGR (downsample $\times$2)             & 0.79 & 0.86 & 0.85 & 0.87 & 0.055 \quad 0.303 \\
\bottomrule
\end{tabular}}
\caption{Ablation of the automated genomic interpretation module.}
\vspace{-9mm}
\label{tab:ablation}
\end{table}
The ablations confirm the necessity of all components.
A strict concept pathway ($\alpha{=}1$) yields the best concept fidelity and calibration,
minimizing expected clinical cost.
Removing concept–fidelity supervision causes the sharpest degradation in mean $R^2$ and AUROC,
showing that the bottleneck must be explicitly guided.
Without prior–consistency, sign errors appear against biological priors and decision stability drops.
Disabling KL alignment lowers both concept and decision AUROCs, evidencing the need for distribution matching.
Without calibration, ECE doubles and over–confident errors directly inflate cost.
Finally, replacing CGR with histogram or low–resolution input consistently reduces separability,
underscoring the importance of the symbolic–to–visual front–end.

\section{Conclusion}
We introduced an automated genomic interpretation module that integrates symbolic-to-visual DNA encoding, interpretable concept bottlenecks, and cost-aware decision policies into a unified framework for medical automation. Unlike conventional black-box classifiers, our design enforces a transparent reasoning path: DNA sequences are transformed into CGR images, encoded via a CNN backbone, constrained through concept fidelity and prior consistency, and finally mapped to calibrated, cost-sensitive recommendations. Experiments across two HIV sequence datasets confirm that the module achieves competitive accuracy, robust concept recovery, and improved utility–retest trade-offs. More importantly, the module provides interpretable evidence and actionable recommendations, establishing a closed loop from sequence input to decision output. This ensures not only predictive accuracy but also auditability, trust, and practical readiness for deployment in clinical and robotic systems. Future directions include scaling the framework to multi-gene panels, integrating richer biological priors, and embedding the module into full-stack medical robots to support autonomous triage, monitoring, and decision support in genomic medicine.

\section*{Acknowledgment}: 
This work is supported by NSF Grants NO. 2340882, 2334624, 2334246, and 2334690.


\bibliographystyle{IEEEtran}
\bibliography{egbib}

@inproceedings{pang2024integrating,
  title={Integrating clinical knowledge into concept bottleneck models},
  author={Pang, Winnie and Ke, Xueyi and Tsutsui, Satoshi and Wen, Bihan},
  booktitle={MICCAI},
  year={2024},
  organization={Springer}
}

@article{kuiken2003hiv,
  title={HIV sequence databases},
  author={Kuiken, Carla and Korber, Bette and Shafer, Robert W},
  journal={AIDS reviews},
  volume={5},
  number={1},
  pages={52},
  year={2003}
}

@inproceedings{yuksekgonul2023posthoc,
  title={Post-hoc Concept Bottleneck Models},
  author={Yuksekgonul, Mert and Bica, Ioana and Zhang, Han and Ghassemi, Marzyeh and Zhang, Michael},
  booktitle={ICML},
  year={2023},
  publisher={PMLR}
}

@inproceedings{zhang2024embodiment,
  title={Embodiment: Self-Supervised Depth Estimation Based on Camera Models},
  author={Zhang, Jinchang and Reddy, Praveen Kumar and Wong, Xue-Iuan and Aloimonos, Yiannis and Lu, Guoyu},
  booktitle={IROS},
  pages={7809--7816},
  year={2024},
  organization={IEEE}
}

@inproceedings{koh2020concept,
  title={Concept bottleneck models},
  author={Koh, Pang Wei and Nguyen, Thao and Tang, Yew Siang and Mussmann, Stephen and Pierson, Emma and Kim, Been and Liang, Percy},
  booktitle={ICML},
  year={2020},
  organization={PMLR}
}

@article{CEM,
  title={Concept embedding models: Beyond the accuracy-explainability trade-off},
  author={Espinosa Zarlenga, Mateo and Barbiero, Pietro and Ciravegna, Gabriele and Marra, Giuseppe and Giannini, Francesco and Diligenti, Michelangelo and Shams, Zohreh and Precioso, Frederic and Melacci, Stefano and Weller, Adrian and others},
  journal={NeurIPS},
  volume={35},
  pages={21400--21413},
  year={2022}
}

@article{SCBM,
  title={Stochastic concept bottleneck models},
  author={Vandenhirtz, Moritz and Laguna, Sonia and Marcinkevi{\v{c}}s, Ri{\v{c}}ards and Vogt, Julia},
  journal={NeurIPS},
  year={2024}
}

@inproceedings{kim2023concept2,
  title={Concept bottleneck with visual concept filtering for explainable medical image classification},
  author={Kim, Injae and Kim, Jongha and Choi, Joonmyung and Kim, Hyunwoo J},
  booktitle={MICCAI},
  pages={225--233},
  year={2023},
  organization={Springer}
}

@inproceedings{yin2021focusing,
  title={Focusing on clinically interpretable features: selective attention regularization for liver biopsy image classification},
  author={Yin, Chong and Liu, Siqi and Shao, Rui and Yuen, Pong C},
  booktitle={MICCAI},
  year={2021},
  organization={Springer}
}

@article{manh2022multi,
  title={Multi-attribute attention network for interpretable diagnosis of thyroid nodules in ultrasound images},
  author={Manh, Van T and Zhou, Jianqiao and Jia, Xiaohong and Lin, Zehui and Xu, Wenwen and Mei, Zihan and Dong, Yijie and Yang, Xin and Huang, Ruobing and Ni, Dong},
  journal={UFFC},
  volume={69},
  number={9},
  pages={2611--2620},
  year={2022},
  publisher={IEEE}
}

@inproceedings{applequist2013brief,
  title={A brief review of recent controversies in the taxonomy and nomenclature of Sambucus nigra sensu lato},
  author={Applequist, Wendy L},
  booktitle={I International Symposium on Elderberry 1061},
  pages={25--33},
  year={2013}
}

@article{lovich2018taxonomy,
  title={Taxonomy: A history of controversy and uncertainty},
  author={Lovich, Jeffrey E and Hart, Kristen},
  year={2018}
}

@article{wang1994complexity,
  title={On the complexity of multiple sequence alignment},
  author={Wang, Lusheng and Jiang, Tao},
  journal={JCB},
  year={1994}
}

@article{zielezinski2017alignment,
  title={Alignment-free sequence comparison: benefits, applications, and tools},
  author={Zielezinski, Andrzej and Vinga, Susana and Almeida, Jonas and Karlowski, Wojciech M},
  journal={GB},
  volume={18},
  number={1},
  pages={186},
  year={2017},
  publisher={Springer}
}

@article{jeffrey1990chaos,
  title={Chaos game representation of gene structure},
  author={Jeffrey, H Joel},
  journal={NAR},
  year={1990},
  publisher={Oxford University Press}
}

@article{lochel2021chaos,
  title={Chaos game representation and its applications in bioinformatics},
  author={L{\"o}chel, Hannah Franziska and Heider, Dominik},
  journal={CSBJ},
  year={2021},
  publisher={Elsevier}
}

@article{kariin1995dinucleotide,
  title={Dinucleotide relative abundance extremes: a genomic signature},
  author={Kariin, Samuel and Burge, Chris},
  journal={Trends in genetics},
  number={7},
  year={1995},
  publisher={Elsevier}
}

@article{randic2013milestones,
  title={Milestones in graphical bioinformatics},
  author={Randi{\'c}, Milan and Novi{\v{c}}, Marjana and Plav{\v{s}}i{\'c}, Dejan},
  journal={IJQC},
  volume={113},
  number={22},
  pages={2413--2446},
  year={2013},
  publisher={Wiley Online Library}
}

@article{hill1992chaos,
  title={Chaos game representation of coding regions of human globin genes and alcohol dehydrogenase genes of phylogenetically divergent species},
  author={Hill, Kathleen A and Schisler, Nicholas J and Singh, Shiva M},
  journal={JME},
  year={1992},
  publisher={Springer}
}

@article{hoang2016numerical,
  title={Numerideep learning approach focal encoding of DNA sequences by chaos game representation with application in similarity comparison},
  author={Hoang, Tung and Yin, Changchuan and Yau, Stephen S-T},
  journal={Genomics},
  volume={108},
  year={2016},
  publisher={Elsevier}
}

@article{han2021comparative,
  title={Comparative analysis and prediction of nucleosome positioning using integrative feature representation and machine learning algorithms},
  author={Han, Guo-Sheng and Li, Qi and Li, Ying},
  journal={BMC bioinformatics},
  year={2021},
  publisher={Springer}
}

@article{lecun1989backpropagation,
  title={Backpropagation applied to handwritten zip code recognition},
  author={LeCun, Yann and Boser, Bernhard and Denker, John S and Henderson, Donnie and Howard, Richard E and Hubbard, Wayne and Jackel, Lawrence D},
  journal={Neural computation},
  year={1989},
  publisher={MIT Press}
}

@inproceedings{safoury2019enriched,
  title={Enriched dna strands classification using cgr images and convolutional neural network},
  author={Safoury, Sarah and Hussein, Walid},
  booktitle={ICBBS},
  year={2019}
}

@article{avila2023accurate,
  title={Accurate and fast clade assignment via deep learning and frequency chaos game representation},
  author={Avila Cartes, Jorge and Anand, Santosh and Ciccolella, Simone and Bonizzoni, Paola and Della Vedova, Gianluca},
  journal={GigaScience},
  year={2023},
  publisher={Oxford University Press}
}

@article{hammad2023hybrid,
  title={A hybrid deep learning approach for COVID-19 detection based on genomic image processing techniques},
  author={Hammad, Muhammed S and Ghoneim, Vidan F and Mabrouk, Mai S and Al-Atabany, Walid I},
  journal={Scientific Reports},
  year={2023},
  publisher={Nature Publishing Group UK London}
}

@article{espinosa2022concept,
  title={Concept embedding models: Beyond the accuracy-explainability trade-off},
  author={Espinosa Zarlenga, Mateo and Barbiero, Pietro and Ciravegna, Gabriele and Marra, Giuseppe and Giannini, Francesco and Diligenti, Michelangelo and Shams, Zohreh and Precioso, Frederic and Melacci, Stefano and Weller, Adrian and others},
  journal={NeurIPS},
  volume={35},
  pages={21400--21413},
  year={2022}
}

@article{lin2025graph,
  title={Graph integrated multimodal concept bottleneck model},
  author={Lin, Jiakai and Zhang, Jinchang and Lu, Guoyu},
  journal={arXiv preprint arXiv:2510.00701},
  year={2025}
}

@inproceedings{zhang2025vision,
  title={Vision-language embodiment for monocular depth estimation},
  author={Zhang, Jinchang and Lu, Guoyu},
  booktitle={CVPR},
  year={2025}
}

@inproceedings{zhang2025exploit,
  title={Exploit your latents: Coarse-grained protein backmapping with latent diffusion models},
  author={Zhang, Rongchao and Huang, Yu and Lou, Yiwei and Xin, Yi and Chen, Haixu and Cao, Yongzhi and Wang, Hanpin},
  booktitle={AAAI},
  year={2025}
}

@article{zhang2024synergistic,
  title={Synergistic attention-guided cascaded graph diffusion model for complementarity determining region synthesis},
  author={Zhang, Rongchao and Huang, Yu and Lou, Yiwei and Ding, Weiping and Cao, Yongzhi and Wang, Hanpin},
  journal={TNNLS},
  year={2024},
  publisher={IEEE}
}

@inproceedings{li2025lion,
  title={Lion-fs: Fast \& slow video-language thinker as online video assistant},
  author={Li, Wei and Hu, Bing and Shao, Rui and Shen, Leyang and Nie, Liqiang},
  booktitle={CVPR},
  year={2025}
}

@inproceedings{li2025cogvla,
  title={Cogvla: Cognition-aligned vision-language-action model via instruction-driven routing \& sparsification},
  author={Li, Wei and Zhang, Renshan and Shao, Rui and He, Jie and Nie, Liqiang},
  booktitle={NeurIPS},
  year={2025}
}

@inproceedings{li2025semanticvla,
  title={Semanticvla: Semantic-aligned sparsification and enhancement for efficient robotic manipulation},
  author={Li, Wei and Zhang, Renshan and Shao, Rui and Fang, Zhijian and Zhou, Kaiwen and Tian, Zhuotao and Nie, Liqiang},
  booktitle={ AAAI},
  year={2026}
}

@article{zeng2025FSDrive,
      title={FutureSightDrive: Thinking Visually with Spatio-Temporal CoT for Autonomous Driving},
      author={Shuang Zeng and Xinyuan Chang and Mengwei Xie and Xinran Liu and Yifan Bai and Zheng Pan and Mu Xu and Xing Wei},
      journal={arXiv preprint arXiv:2505.17685},
      year={2025}
      }

@article{zhang2025adaptive,
  title={Adaptive Event Stream Slicing for Open-Vocabulary Event-Based Object Detection via Vision-Language Knowledge Distillation},
  author={Zhang, Jinchang and Li, Zijun and Lin, Jiakai and Lu, Guoyu},
  journal={arXiv preprint arXiv:2510.00681},
  year={2025}
}

@inproceedings{zhang2024underground,
  title={Underground mapping and localization based on ground-penetrating radar},
  author={Zhang, Jinchang and Lu, Guoyu},
  booktitle={ACCV},
  year={2024}
}

\end{document}